\documentclass[sigconf]{acmart}
\AtBeginDocument{%
  }

\copyrightyear{2026}
\acmYear{2026}
\setcopyright{cc}
\setcctype{by}
\acmConference[FDG '26]{Foundations of Digital Games}{August 10--13, 2026}{Copenhagen, Denmark}
\acmBooktitle{Foundations of Digital Games (FDG '26), August 10--13, 2026, Copenhagen, Denmark}
\acmDOI{10.1145/3815598.3815619}
\acmISBN{979-8-4007-2495-4/2026/08}

\acmISBN{978-1-4503-XXXX-X/2018/06}

\usepackage{algorithmic}
\usepackage{algorithm2e}

\begin{document}

\title{Representing and Generating Levels Over Time through Playtrace Reconstructive Partitioning}


\author{Emily Halina}
\affiliation{%
  \institution{Department of Computing Science, Alberta Machine Intelligence Institute (Amii), University of Alberta}
  \city{Edmonton}
  \state{Alberta}
  \country{Canada}
}
\email{ehalina@ualberta.ca}

\author{Matthew Guzdial}
\affiliation{%
  \institution{Department of Computing Science, Alberta Machine Intelligence Institute (Amii), University of Alberta}
  \city{Edmonton}
  \state{Alberta}
  \country{Canada}
}
\email{guzdial@ualberta.ca}


\begin{abstract}
Video games are a dynamic medium experienced over time.
While there are many Procedural Content Generation (PCG) approaches for generating video game levels, they often use representations that abstract away this dynamic nature.
In this paper, we introduce a novel, domain-independent ``cake'' representation for game levels over time which implicitly encodes dynamic information.
We present a novel level generation approach Playtrace Reconstructive Partitioning (PRP) specifically developed for this cake representation.
We compare against six state-of-the-art PCG approaches in the game domain of \textit{Sokoban}, and find that our approach can generate valid levels without sacrificing solution diversity.
We believe our cake representation more neatly encodes the implicit dynamic nature of games compared to existing representations, which allows for our domain-agnostic level generation algorithm PRP. 
\end{abstract}


\maketitle

\section{Introduction}


Since the beginning of technical games research, there has been a question of how to represent video game levels.
Common solutions include tile and graph-based representations \cite{summerville2016vglc}, memory \cite{mawhorter2021content}, and images \cite{snodgrass2018towards,mirgati2023joint}.
All of these representations share a similarity: they treat levels as if they are static.
However, one of the fundamental features of video games is that they are dynamic: entities move and transform over time according to player input.
This means without an external source of dynamic information, it is impossible to guarantee that new levels have desired qualities, such as playability. 

There have been many prior approaches for addressing the discrepancy between static level representation and dynamic gameplay.
In the commonly used tile representation of \textit{Super Mario Bros.} from the Video Game Level Corpus (VGLC) \cite{summerville2016vglc} the player path is encoded as a separate token onto an existing tile representation \cite{summerville2016a}.
While this can help bias a generator towards playability, this representation is limited by its two-dimensional nature.
For example, if the player backtracks to the same tile multiple times, it becomes impossible to represent the true path.
We can also encode dynamic information through hand-authored or learned constraint information \cite{cooper2022sturgeon,vandara2025spacetime}, or by including an automated solver in a reward or evaluation function \cite{khalifa2020pcgrl}.
However, this does not change the level representation, instead approximating this dynamic information elsewhere in the generation process.
These approaches require the manual encoding of design information into constraints or functions, which could be burdensome or time-consuming for developers, particularly for a new game domain.
If we could better address the discrepancy between existing static level representations and the true dynamic gameplay of video games, we could potentially improve generalizability across tasks such as level generation and automated game playing.
This could aid game designers aiming to implement these systems in their own games.


In this paper, we formally define the \textbf{cake} representation for representing levels over time, naturally encoding dynamic gameplay information.
A cake represents a playtrace of a level as a series of $k$ discrete timesteps, where the global game state is recorded at each timestep.
These ``slices'' of game state represent the level over time, hence cake.
While similar temporal representations appear in prior work \cite{vandara2025spacetime}, the cake representation is distinct in that each ``temporal entity'' is tracked separately over time, allowing for the automated encoding of game mechanics without separate hand-authoring.
To showcase the strengths of the cake representation, we introduce a new method for level generation, \textbf{Playtrace Reconstructive Partitioning} (PRP) which takes advantage of the cake representation to generate valid levels.
Inspired by Tree-based Reconstructive Partitioning \cite{halina2023tree}, PRP works by performing an adapted binary space partitioning across time over the cake representation, matching dynamic entities to generate both a level and its solution simultaneously.
We compare against six state-of-the-art level generation approaches for \textit{Sokoban}, and find that PRP performed equivalently or better than these baselines in terms of playability and solution diversity without explicitly encoded constraints or reward signals.

The contributions of this paper are as follows:
\begin{enumerate}
    \item The cake representation, a novel representation for game levels as slices of game state over time.
    \item PRP, an algorithm which uses the cake representation to generate valid levels and their solutions simultaneously.
    \item An evaluation of PRP in the game domain of \textit{Sokoban} against six state-of-the-art baselines. 
    \item An ablation study of PRP over different possible inputs, representing different potential levels of designer control.
    \item Two case studies, one in an existing domain (\textit{OvercookedAI}) and one in a novel game domain (\textit{SkeleWalker}).
\end{enumerate}

\section{Related Work}
In this section, we discuss work on representing levels, and overview PCG approaches applied to \textit{Sokoban}, our evaluation domain.

\subsection{Level Representation}


There exist many static level representations, including tiles \cite{summerville2016vglc}, graphs \cite{kim2019automatic}, images \cite{mirgati2023joint}, and memory \cite{mawhorter2021content}.
Several of these representations explicitly encode dynamic information via affordances.
For example, the VGLC includes a ``breakable'' affordance which encodes the notion that a tile can break as in \textit{Super Mario Bros.} \cite{summerville2016vglc}.
Other approaches rely on the representation itself, as in tile embeddings \cite{jadhav2021tile} and memory-based representations \cite{mawhorter2021content}. 
Still others use external rules or functions which act on the representation but are not encoded within it, such as constraints or re-write rules \cite{cooper2023sturgeon}.
However, generally the level representation does not include sufficient information to fully encode a potential playthrough. 


Some prior level generation work has attempted to use videos as training data for machine learning-based approaches in an attempt to encode dynamic information about the game state \cite{guzdial2016game,mirgati2023joint}.
However, these approaches do not generally consider dynamic information, instead representing each frame individually as an image.
World models, learned representations of an existing environment, implicitly encode level representation as pathways through the learned latent space \cite{ha2018world}.
While this representation is dynamic, there is no way at present to disentangle the level information from the representation of the game itself.

A notable exception to the static representations discussed above is the representation defined by \citeauthor{vandara2025spacetime} in their work on Spacetime Level Generation using Sturgeon, an existing family of constraint-based level generation approaches (Sturgeon-ST) \cite{vandara2025spacetime}.
Vandara et al. proposed a modification to Sturgeon over three-dimensional space-time blocks encoded as two-dimensional film strips that represent a level and its solution over time.
They do not formally define their film strip representation, so we cannot clearly differentiate it from our cake representation (defined below). 
However, we note that their implementation does not track temporal entities, which form the bedrock of our representation. 
Another major difference comes from the generation approach, which only considers local temporal constraints, meaning it can fail to respect global temporal constraints like there only being a single player entity. 
Vandara et al. addressed this through the addition of hand-authored constraints for the first frame of generated gameplay \cite{vandara2025spacetime}.
As Sturgeon-ST is the most similar prior approach we include it as a baseline in our evaluation.

\begin{figure*}[th]
    \centering
    \includegraphics[width=\linewidth]{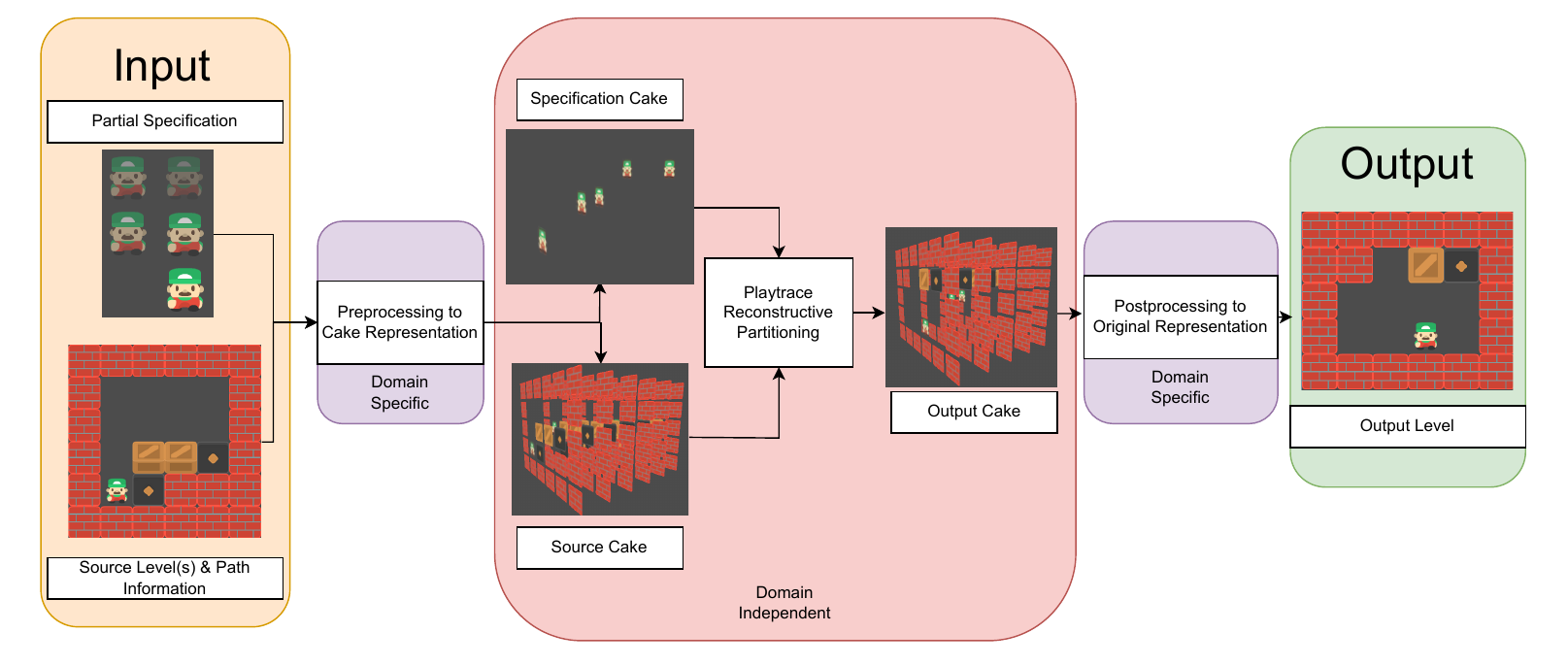}
    \caption{Visualization of the PRP level generation pipeline. As input the system requires a partial specification alongside a corpus of one or more source levels. After processing the input into our cake representation, we perform PRP to assemble the output cake, then perform postprocessing to retrieve the output level. \textit{Sokoban} graphics by Kenney.}
    \label{system-overview-figure}
\end{figure*}

\subsection{PCG for \textit{Sokoban}}
\textit{Sokoban} is a well-explored game domain for level generation \cite{zakaria2022procedural}.
We identify two categories of prior approaches: those reliant on training data, and approaches reliant on human-authored functions or other external information.
Our approach falls into the former category, as we draw the majority of the information required to generate levels from the cake representations derived from example playtraces of human-authored levels.

Level generation approaches that use training data fall under the category of PCG via Machine Learning (PCGML) \cite{summerville2018procedural}.
Numerous PCGML approaches have been applied to \textit{Sokoban} \cite{earle2021learning, siper2022path}.
A subset of these approaches are neural network-based level generation approaches, which use various neural network architectures to model the level generation process \cite{suleman2017generation}.
Zakaria et al.'s survey paper on PCGML for \textit{Sokoban} level generation is a foundational work comparing many of these existing approaches \cite{zakaria2022procedural}.
As such, our evaluation is largely informed by their work.

While some neural network-based approaches have found success in \textit{Sokoban}, the domain can prove challenging without additional external information due to its strict global coherency constraints \cite{zakaria2022procedural}.
We believe that our cake representation may be helpful to these approaches, as the representation contains more information on a game's dynamics.
As a point of comparison, we use multiple neural network based approaches as baselines, including Path of Destruction \cite{siper2022path} and Long Short-Term Memory Recurrent Neural Networks (LSTM) \cite{suleman2017generation}.

The second category of approaches do not require training data for generation.
Instead, these approaches rely on human-authored external signals such as functions or constraints.
For example, search-based PCG approaches use a human-authored fitness function to generate levels and have been applied to \textit{Sokoban} \cite{schaa2021generating}.
However, these approaches can suffer from slow inference speeds and convergence issues.
A natural extension of search-based methods is the PCG via Reinforcement Learning (PCGRL) approach, in which an agent trains to make iterative changes to a level guided by an authored reward function \cite{khalifa2020pcgrl}.
While this approach requires a domain-specific hand authored reward function to guide the training process, Zakaria et al. found it to be highly performant in the \textit{Sokoban} domain \cite{zakaria2022procedural}.
As such, we include two top performing PCGRL settings as baselines.

Constraint-based methods use authored constraints or re-write rules to generate levels \cite{beukman2023hierarchical, cooper2022sturgeon}.
While these approaches can guarantee global coherency by enforcing playability constraints, these constraints require expert domain knowledge to author.
In comparison, our approach requires relatively little hand-authored information outside of processing to the cake representation. 
We use the constraint-based method Sturgeon-MKIII as a baseline \cite{cooper2022sturgeon}, as it is a modern example of a constraint-based method.

\section{System Overview}

In this section, we discuss our cake representation and Playtrace Reconstructive Partitioning (PRP). 
Figure \ref{system-overview-figure} depicts a visualization of the PRP level generation pipeline based on existing level data, using \textit{Sokoban} for ease of visualization.
As input, the generation process requires two sources of dynamic information.
The first is a corpus of source levels with corresponding playtrace data from which to draw temporal entities.
The second is a partial specification for the output level, which may be empty.
Both inputs are processed into \textbf{cake} representations that encode the complete state of the level at every timestep as slices over time.
We note that in practice, the source levels may be processed into the cake representation and stored, particularly in domains where the processing may be computationally expensive.
PRP takes these cake representations as input, and matches objects from the source levels into a partially specified output via a Binary Space Partitioning (BSP) over time.
BSP is an existing level generation approach that partitions spaces from existing levels to fill in sections of a new level \cite{snodgrass2019levels, halina2023tree}.
PRP yields an output cake which is processed into the final output level via a domain-specific post-processing step.

\subsection{Input \& Requirements}
Our generation pipeline requires two sources of dynamic information: a partial specification of the output level, and a corpus of one or more source levels with playtrace information.
The partial specification partially specifies the output level, constraining our pattern-matching process.
This partial specification can include an arbitrary amount of detail to allow for flexible designer control of the output levels.
For example, a valid partial specification could be a player path or entities defining structure such as walls. 
The corpus of source levels provides the temporal entities used by PRP to fill in the partial specification of the output level.
For each source level in the corpus, PRP requires a playtrace of said level which the algorithm uses to inform its generation.
We make a number of assumptions, outlined in the next subsection, based on this playtrace which makes PRP a domain-agnostic level generation algorithm.
We argue that this allows PRP to better support designers of a novel game.


\subsection{Cake Representation}

\begin{figure}
    \centering
    \includegraphics[width=\linewidth]{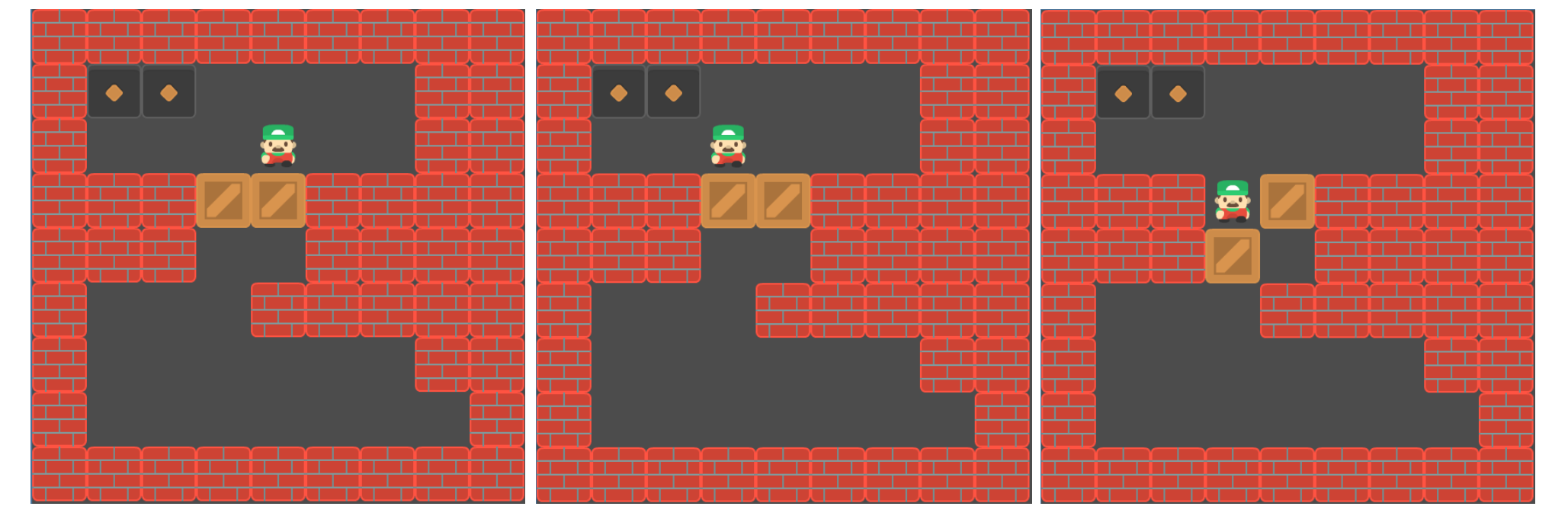}
    \caption{A visualized segment of a cake representation of a playtrace for a human-authored level. Note that the cake representation for this paper uses a character-based tile format. Each game state is separated by a delimiter token to denote beginning and end points.}
    \label{fig:cake figure}
\end{figure}

Figure \ref{fig:cake figure} depicts a segment of a cake representation for an example \textit{Sokoban} level playtrace.
We define a cake representation for a given game level as a $k \times n \times m$ tensor representing a Markovian sequence of $n \times m$ game states over $k$ timesteps.
Within the cake representation, we define temporal entities as distinct entities of a type $T$ which persist between timesteps.
For example, a temporal entity in \textit{Sokoban} could be the player, or an individual box, with their types being player and box respectively.
We note that in some cases, like the walls in \textit{Sokoban}, it may be useful to consider a temporal entity that takes up more than one spatial position. 

Our representation makes three assumptions.
The first is that the sequence of states is Markovian, meaning that each state contains all the information necessary to predict the next state independently.
This is necessary, as otherwise the representation would not encode the dynamic information of the game.
The second is that the representation must delineate between types of temporal entities over time.
The token-based encoding of a game level mostly covers this assumption, and is a standard approach for most data-based level generation.
The third is that it must track temporal entities over time, meaning we can tell when an entity has moved to a different location or transformed from one type to another.
For example, in \textit{Sokoban} we have a box type which transforms into a solved box when placed on a goal.
Notably, a solved box can transform back into a box if pushed from its current position, which is required for certain solutions.
This third assumption encodes domain-specific knowledge, 
but this knowledge can be learned \cite{guzdial2017game}. 

\subsection{Playtrace Reconstructive Partitioning}

\begin{algorithm}[t]
\caption{Playtrace Reconstructive Partitioning}
\textbf{Input:} $k_p \times n_p \times m_p$ partial specification cake $P$, set of source cakes $S$\\
initialize output $O$ as empty $k_p \times n_p \times m_p$ tensor\\
\For{$\textup{temporal entity } p \in P$}{populate $O$ with $p$} 
\For{type $T \in S$ \textup{in order of} SelectionPolicy()}{
\For{\textup{temporal entity} $s \in S$ \textup{of type} $T$}{\textit{matches} = $\{\}$\\\For{\textit{pos} $ = (z, x, y) \in O$}{
    \If{MatchesAt($s$, pos)}{add \textit{pos} to \textit{matches}}}\If{matches $\neq$ $\{\}$}{\textit{pos} $\gets$ sample from \textit{matches}\\
    populate $O$ with $s$ at \textit{pos}}}}

    
   \textbf{Return:} output cake $O$
\end{algorithm}

Algorithm 1 describes Playtrace Reconstructive Partitioning (PRP).
This algorithm relies on two sources of domain knowledge, encoded by the Selection Policy and MatchesAt functions, which we describe in detail below. 
The algorithm takes in two sources of input: a partial specification processed into the cake representation $P$, and a set of cake representations of source levels $S$ from which to draw temporal entities.
To begin, we initialize the output as specified by the dimensions and content of the partial specification cake.
This process involves creating an empty $k_p \times n_p \times m_p$ tensor $O$, then populating the tensor with each of the temporal entities present in the partial specification.
If the partial specification is empty, the dimensions of $O$ are determined by temporal entity selection.

The first piece of domain knowledge required by PRP takes the form of the Selection Policy function, which returns an ordering over the types of temporal entities present in a cake representation. 
This is domain-specific as a particular set of types are only shared by a particular game domain. 
For example, for \textit{Sokoban} we have the player, the boxes, solved boxes, goals, walls, and empty tiles. 
We note that this ordering could be arbitrary but in practice it makes sense to place the types that most constrain the output earlier in this ordering.
In particular, in \textit{Sokoban} we used an ordering of players, boxes, and then all other types tied for third. 

We iterate through the types $T$ of temporal entities based on the Selection Policy.
For each type $T$, we then iterate through all temporal entities $s$ of type $T$ in $S$. 
For each temporal entity $s$ we attempt to match it to $O$ using the MatchesAt function. 
This is a domain-specific matching function, which indicates if a temporal entity $s$ from $S$ can exist in a particular position in time in $O$. 
These represent global coherency constraints, similar to the ones used by Sturgeon-ST \cite{vandara2025spacetime}.
For each temporal entity $s$, if it matches with the current output cake $O$ we add it to a set of matches on a per-entity basis. 
If there are multiple matches we uniformly sample from them to place the entity $s$ in the output cake $O$ at a particular position in time.
The process continues until we run out of temporal entities to add and types of temporal entities. 
There is one special case in this function, which is that if a temporal entity $s$ is stationary spatially throughout all timesteps and occupies multiple positions, we run BSP on it, allowing us to break the temporal entity apart along the space axes if necessary for matching.


As a practical example in \textit{Sokoban}, consider a case with a partial specification of just a player path. 
From there, the algorithm would attempt to add other players to the output Cake $O$ and fail due to the MatchesAt global constraint of a single player. 
Next, it would attempt to add in as many boxes as possible, where the boxes' paths (their temporal information) matches with the player path. 
In other words, the player could have pushed these boxes by walking along the provided player path. 
From there, all other entities are added until the output cake has an entity in each position.
After this process, we return the output cake, a representation that captures both the generated level and an associated playtrace.

There is one additional case not covered by the above pseudocode, which is that the generation process can fail. 
Failure in this case means an output cake that includes null tokens, i.e., not all positions have been filled with temporal entities. 
For example, this can happen in the case when a selected player path does not match with any other information from $S$. 
As the null token is domain-agnostic, this failure check is also domain-agnostic, though additional failure cases may be added based on a domain. 

\subsection{Postprocessing Output Cake}
After PRP completes we have an output cake containing $k$ frames of level information across time.
To transform this representation into the final level we could simply take the first slice of the cake, which represents the initial state of the generated level.
In the case of this paper, we perform this simple post-processing step.
However, there are many other possible ways to use the additional information in the cake representation when constructing the final output level.
For example, in a game like \textit{Sokoban}, we could automatically take the level at a later, partially solved state, and return this as an ``easier'' version of the same level.

\subsection{\textit{Sokoban} Implementation} 
This subsection outlines the details and justifications of our design choices for the pre-processing and post-processing steps of our \textit{Sokoban} PRP implementation, as well as the three aspects of domain-specific knowledge: the cake representation processing, the Selection Policy, and the MatchesAt function. 

For pre-processing, as the \textit{Sokoban} input levels were already on a tile-based grid, we were able to perform a one-to-one transfer of each tile type to a text representation.
This text representation directly maps each tile type to a specific text character.
We used an existing dataset of twelve $7 \times 7$ human authored source levels for our approach which has been used in prior work for evaluating existing PCG approaches for \textit{Sokoban} \cite{zakaria2022procedural}.
We ran a breadth-first search solver to collect playtrace information, represented in the form of a series of inputs.
Notably, this solver is \textbf{not} a required component of PRP, and was simply used to find the shortest possible solution for each \textit{Sokoban} level.
PRP can generate levels with playtraces from arbitrary sources, such as human playthrough data. 
We demonstrate this in our case studies below.
We used the shortest possible solution to avoid biasing the generator to create more complex levels through using longer than necessary input paths, and to reduce computation time when solving each level.

We processed the playtraces into the cake representation by a custom implementation of \textit{Sokoban}'s logic, with each player move representing a new slice in each cake.
The text-based representation of each source level, alongside their corresponding cake representations, can be found in the project's Github repository\footnote{https://github.com/emily-halina/PRP-Sokoban}.
For post-processing, we took the first slice of each output cake created by PRP if it contained boxes.
If it did not contain boxes, we considered the level trivial and discarded it. 

For the Selection Policy for temporal entities we order in terms of dynamism.
In \textit{Sokoban}, the player is inherently the most dynamic entity, as it moves each timestep, and as such we order the player type first.
The next most dynamic entity type is the box, which we order second, followed by all non-moving entities.

For the MatchesAt function, we specified three constraints.
The first was that only one player token was allowed at any given timestep.
The second is that no player, box, or wall could occupy the same space at the same timestep.
The third is that no part of the temporal entity went out-of-bounds of the dimensions of the output cake $O$.

When performing generation for \textit{Sokoban}, we discarded trivial generated levels which contained no boxes.
We chose to consider these levels as failed generations as they are by default solved.
We report approximate generation times and failure rates in the Ablation subsection of the Evaluation section.

\section{Evaluation}

In this section we overview the evaluation of PRP in the domain of \textit{Sokoban}.
To evaluate our approach, we compared against six baselines.
These baselines were Path of Destruction (PoD) \cite{siper2022path}, Sturgeon-MKIII \cite{cooper2022sturgeon}, Sturgeon-ST \cite{vandara2025spacetime}, two PCGRL implementations \cite{khalifa2020pcgrl}, and an LSTM-based recurrent neural network approach \cite{zakaria2022procedural}.
The implementation details and parameters for each baseline are covered in the Baselines subsection.


The majority of our evaluation is based on Zakaria et al.'s survey paper on deep learning approaches for \textit{Sokoban} \cite{zakaria2022procedural}.
We chose to base our evaluation on this work as it represents the most up-to-date and rigorous study of level generation approaches for \textit{Sokoban}.
As such, we report results directly from their paper for the PCGRL and LSTM approaches, and adapt their metrics for comparison purposes.
To supplement these results, we draw directly on corpora of levels from prior work for PoD and Sturgeon, which are approaches not evaluated by Zakaria et al.
We chose to use existing corpora rather than re-implementing these approaches and regenerating populations of levels to maintain consistency on which future work can build.
We ran Sturgeon-ST with minor implementation adjustments on our training levels as the original Sturgeon-ST only gave the outputs for a single Sokoban training level \cite{vandara2025spacetime}.

Alongside our baseline evaluation, we wanted to evaluate the effects of changing the partial specification provided to our approach.
We therefore identified three partial specifications and performed an ablation study over them.

\subsection{Baselines}
\subsubsection{Path of Destruction (PoD)} Path of Destruction (PoD) is a level generation approach that iteratively translates random noise into a playable level in a process similar to diffusion models \cite{siper2022path}.
At each timestep, a repair network takes the current tile position and level state and returns an edit to make at the given position.
The current tile position is then updated, whether at random or sequentially, and the process repeats.
This repair network is trained by iteratively ``destroying'' solvable levels, hence the name.
We sampled 1000 generated levels from an existing corpus of solvable PoD generated levels provided by the authors of \cite{siper2022path}, since the loop in theory does not end until the level is solvable.
The levels in the corpus were generated with an observation and goal size of five. 
PoD does not directly encode any dynamic information in its model, but this implementation had an additional external check for solvability before generation ended, representing a distinct way of authoring dynamic information into the generation process.

\subsubsection{Sturgeon-MKIII} Sturgeon is a group of level generation approaches designed to enable constraint-based level generation \cite{cooper2022sturgeon}.
In particular, Sturgeon-MKIII generates both a level and a playtrace simultaneously through tile re-write rules and constraints.
These tile re-write rules are a hand-authored representation of the game's dynamics. 
Thus this baseline represents an alternative way of representing dynamic information in the level generation process.
We sampled 1000 generated \textit{Sokoban} levels from the ``solvable'' section of the Generated Game Level Corpus (GGLC) \cite{bazzaz2025analysis}, which were generated with these tile-rewrite rules \cite{cooper2023sturgeon}.
These levels were generated with solvability as a constraint, and each level contains two boxes and two goals.
Notably, these levels had a size of $8 \times 8$ rather than the $7 \times 7$ of the other approaches.
While this limitation may lead to a slight difference in the number of unique solutions present in each level, this is strictly an advantage for Sturgeon in terms of population diversity.
This is because there are more possible levels in an $8 \times 8$ grid.

\subsubsection{Sturgeon Spacetime (Sturgeon-ST)}
Sturgeon-ST is a level generation approach which modifies Sturgeon to act over three-dimensional space-time blocks encoded as two-dimensional film strips \cite{vandara2025spacetime}.
As Sturgeon supports global constraints, Stugeon-ST introduces additional hand-authored constraints to allow for more controllability. 
Because the implementation of Sturgeon-ST we had access to could only take in a single level as input at a time, we generated 84 levels from each training level to get as close to 1000 as possible, yielding 1008 generated levels.
For each generated level, we specified that there can be only one player token in the first frame to ensure playability, and that the number of boxes in the first frame must equal the number of boxes found in the respective training level.
This constraint was a necessary addition, as without it the approach generated levels without any boxes. 
We also specified each level to have a maximum solution length of 20, though notably this was an upper bound, and some generated levels had shorter solutions.
We chose this solution length as it was the solution length used in the author's \textit{Sokoban} case study, and represented an upper bound on the solution lengths present in the training data.
We included Sturgeon-ST as it is the most related prior work, similarly defining levels over time, but reliant on Sturgeon for generation and with additional authored global constraints.

\subsubsection{PCG via Reinforcement Learning (PCGRL)}
PCG via Reinforcement Learning (PCGRL) is a level generation approach that models the generation process as a Markov Decision Process (MDP), then attempts to learn an agent to solve the MDP.
The goal of the learned agent is to maximize the sum of expected rewards while interacting in the environment.
The reward of a given level state is in part based on the level's path length and solvability, which is assessed by an external BFS solver.
Thus this represents another distinct approach for representing dynamic information, not explicitly in the level generation process but while training the level generator.

For PCGRL, we directly report the results from \cite{zakaria2022procedural} for comparison, which evaluated 10000 generated levels for each PCGRL condition.
In particular, we selected two conditions to report: Turtle and Wide.
The difference between the two conditions is the form of the observation and action set for the agent.
In the Turtle condition, the agent has a position in the level, and take actions to move locations orthogonally or edit the tile at its specified location.
In the Wide condition, the agent can edit any tile freely, and takes actions in the form of (position, tile to edit).
We chose these conditions as they performed best in terms of diversity and playability among the PCGRL conditions reported by Zakaria et al.~\cite{zakaria2022procedural}.

\subsubsection{Long Short-Term Memory (LSTM)} 
Long Short-Term Memory Recurrent Neural Networks (LSTM) are a type of neural network applied to level generation.
On a high level, LSTMs applied to level generation work by generating tiles sequentially, commonly in a ``snake''-like pattern \cite{summerville2016a}.
As the name suggests, the defining characteristic of LSTMs is the memory unit, which holds pieces of input in memory for a learned amount of time.
This helps LSTMs generate content with sequential structure, such as game levels.

For the LSTM generator, as with the PCGRL approaches, we directly report the results from \cite{zakaria2022procedural} over 10000 generated levels.
In particular, we report on the condition which includes sampling level conditions from a Gaussian Mixture Model (GMM).
We chose to report this condition as it performed best among the LSTM variants evaluated by Zakaria et al.
In their implementation, the network had two LSTM layers with a hidden size of 128 each, and a fully connected layer for output.
For more details regarding the implementation of the network and the GMM condition sampling, see Zakaria et al.'s paper \cite{zakaria2022procedural}.
We included an LSTM as it represents a common PCGML approach that does not encode any dynamic information.

\begin{table*}[t]
\begin{tabular}{c|ccccc}
                               & \textbf{Playability $\uparrow$} & \textbf{Duplicates $\downarrow$} & \textbf{Tile Diversity $\uparrow$} & \textbf{Unique Solutions $\uparrow$} & \textbf{Unique Signatures $\uparrow$} \\ \hline
\textbf{Human}         & 100\%       & 0.00\%     & 36.77\%                 & 2.083                     & 0.917                      \\ \hline
\textbf{PoD}                   & \textbf{100\%}       & 71.98\%             & 15.48\%                 & 0.508                     & 0.272                      \\
\textbf{Sturgeon-MKIII (8x8)}        & \textbf{100\%}       & 1.20\%              & 29.43\%                 & \textbf{2.138}            & 0.122                      \\
\textbf{Sturgeon-ST} & \textbf{100\%} & 18.35\% & 32.25\% & 0.748 & 0.077\\ \hline
\textbf{PCGRL Turtle ($\sim$) \textdagger} & 86.40\%              & \textbf{0.00\%}     & \textbf{53.00\%}        & 0.864                     & 0.624                     \\
\textbf{PCGRL Wide ($\sim$) \textdagger}   & 86.60\%              & \textbf{0.00\%}     & 26.00\%                 & 0.818                     & \textbf{0.651}            \\
\textbf{LSTM (w/ GMM) \textdagger}         & 73.4\% $\pm$ 5.3\%      & 10.00\% $\pm$ 6.7\%    & 43.00\% $\pm$ 3.0\%        & 0.315 $\pm$ 0.006            & 0.073 $\pm$ 0.017      \\ \hline
\textbf{PRP (ours)}         & \textbf{100\%}       & \textbf{0.00\%}     & 50.86\%                 & 1.924                     & 0.534         
\end{tabular}
\caption{Comparison of PRP against our baseline approaches across our metrics. We group the approaches in terms of the original Human training levels, the approaches where we collected the metrics ourselves, the reported metrics, and our own approach. We normalize the unique solutions and signatures based on the number of levels generated by each approach. The highest values in each column are bolded, excluding the human dataset results. Baseline values taken from \cite{zakaria2022procedural} are denoted by \textdagger. ($\sim$) denotes the approach was run using a stochastic policy. The LSTM values are averaged across five training runs.}
\label{results-table}

\end{table*}

\begin{table*}[]
\centering
\begin{tabular}{c|ccccc}
\textbf{}              & \textbf{Playability $\uparrow$} & \textbf{Duplicates $\downarrow$} & \textbf{Tile Diversity $\uparrow$} & \textbf{Unique Solutions $\uparrow$} & \textbf{Unique Signatures $\uparrow$} \\ \hline
\textbf{PRP (dataset)} & 100\%                & 0.00\%              & 50.86\%                 & \textbf{1.924}            & \textbf{0.534}             \\
\textbf{PRP (path)}    & 100\%                & 0.00\%              & \textbf{53.00\%}        & 1.347                     & 0.264                      \\
\textbf{PRP (border)}  & 100\%                & \textbf{27.60\%}    & 23.58\%                 & 0.257                     & 0.094                     
\end{tabular}
\caption{Ablation study of PRP conditions based on altering the input partial specification. The highest values in each column are bolded.}
\label{ablation-table}
\end{table*}

\subsection{Metrics}
To evaluate the performance of each generator we aimed to examine two aspects of a level's quality: playability and diversity. Notably, these metrics were adapted from \cite{zakaria2022procedural} to ensure consistency between their results and ours, though some had to be reimplemented according to the descriptions in their paper.

We define playability as the level being possible to complete, i.e., there is some sequence of inputs such that the level reaches a solved state (all boxes are solved boxes).
To measure playability, we implemented a Breadth-First Search (BFS) solver which exhaustively visited every unique state of each level until all solutions were found.
Notably, this solver was only used for computing the evaluation metrics, \textbf{not} as part of the PRP level generation pipeline.
We consider a level playable if the solver finds at least one solution.
We consider a solution unique if every state encountered along the path to the solution was also unique, discounting solutions which include cycles of states.
The solver logged each encountered solution in order of length, which we use for our diversity metrics.

\begin{figure}[t]
    \centering
    \includegraphics[width=0.9\linewidth]{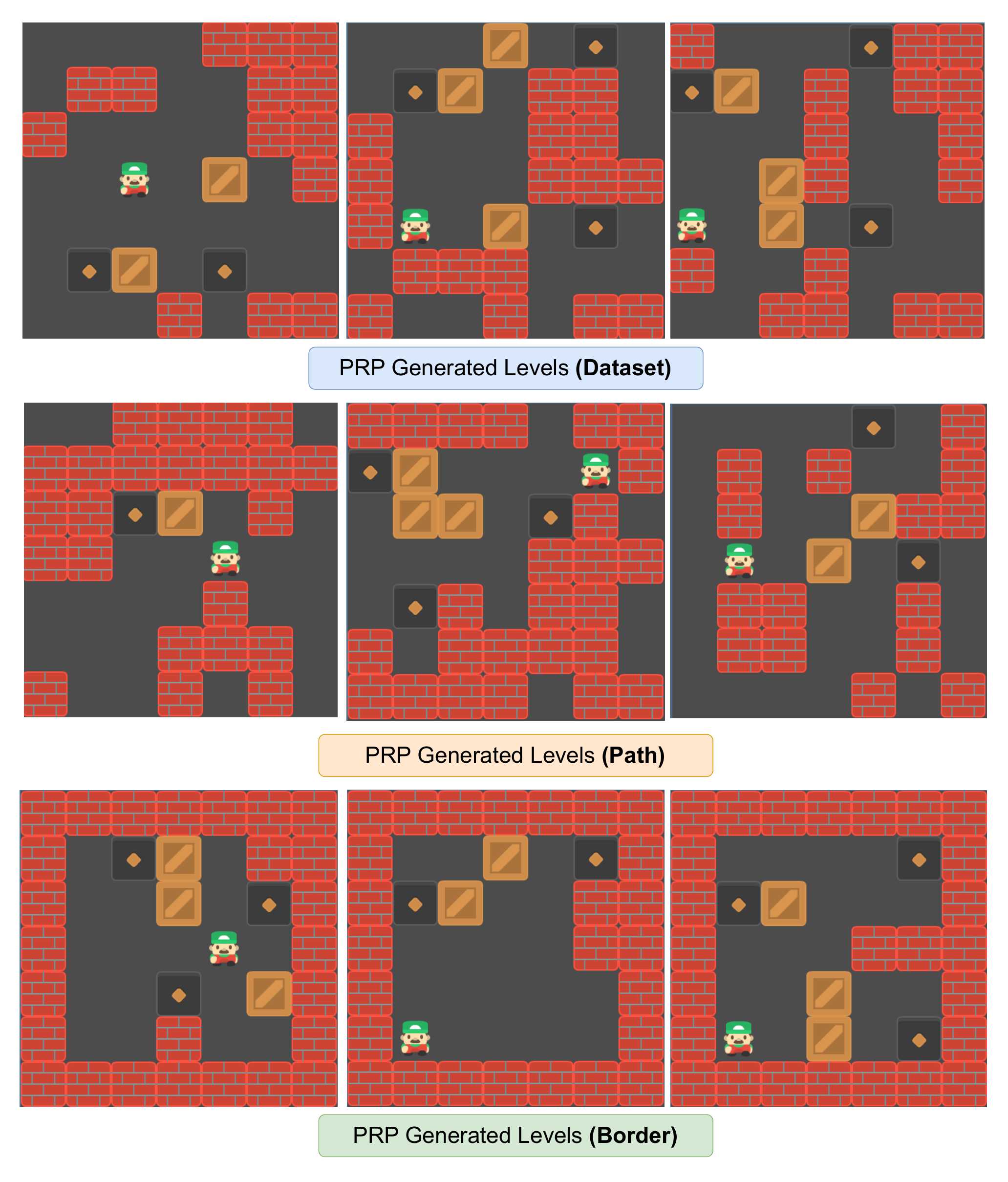}
    \caption{Three randomly selected generated PRP levels from each condition.}
    \label{fig:sokoban levels}
\end{figure}

Diversity refers to both the aesthetic differences between levels in a population, as well as the differences between their solutions.
To capture both of these aspects of diversity, we used four different metrics.
The first is \textbf{Duplicates}, which simply counts the duplicate levels found within a population.
This metric identifies the rate at which a generator produces unique levels.
The second is \textbf{Tile Diversity}, which measures the percentage of tiles which differ within the population of levels on average.
We measure this percentage by calculating the Hamming distance, or edit-distance between each unique pair of levels in the population.
The third is \textbf{Unique Solutions}, which measures the rate of unique solutions found within a population, which we divided by the total number of levels in each population.
This gives a cursory measure of how ``unique'' generated levels are compared to one another, but is an imperfect way of measuring uniqueness on its own.
An example of a potential problem with measuring solution diversity using only the unique solution count is that two solutions can take very different paths, but ultimately end up pushing the boxes in the same directions to solve the level.
To get around this issue, we include our last metric \textbf{Unique Signatures}, in which solutions are generalized into a rotation and flip invariant representation which condenses the solution path to only the directions in which boxes are pushed.
For more details on the implementation of unique signatures and the justification behind using this metric, see Zakaria et al.'s paper~\cite{zakaria2022procedural}.

\subsection{Ablation}

We wanted to investigate the effects of varying the partial specification as a source of input for our approach.
Intuitively, varying the partial specification could drastically change the population of output levels, allowing for greater designer control. 
However, it could come at the cost of decreasing level diversity or impacting playability.
As such, we performed an ablation study across three settings for PRP.
A visual example of each of the three settings can be found in Figure \ref{fig:sokoban levels}.

The first of these settings was the \textbf{dataset} setting, in which the provided partial specification was left empty.
In this case, PRP draws all temporal entities in each output level from the dataset, hence the name.
We performed generation 1000 times as specified in the System Overview section, giving us a total of 1000 generated levels.
This generation process took 24.804 seconds, averaging out to roughly 0.025 seconds to generate a single level, with approximately 140 re-generations required due to trivial generations.

The second setting was the \textbf{path} setting, in which we used a player path as the partial specification.
These player paths ranged in length between ten and twenty inputs which were selected uniformly at random without regard for repeats or cycles.
Specifically, we generated ten randomized paths of each path length between ten and twenty, and generated ten levels based on each of those paths, giving us a total of 1000 generated levels.
This process took 16.812 seconds, averaging out to roughly 0.017 seconds on average to generate a single level, with approximately 230 re-generations required due to trivial generations.
This is slightly faster than the dataset condition, as we do not need to iterate through all of the potential player entities to populate into the level, and can just use the provided path.

The third setting was the \textbf{border} setting, in which we populated the partial specification with a border of wall tokens surrounding the level at each timestep.
This allowed for the exclusive generation of boxed in levels.
Notably, not all of the levels in the human authored dataset we used for generation follow this pattern.
We performed generation 1000 times, giving us a total of 1000 generated levels.
This process took 31.241 seconds, averaging out to roughly 0.031 seconds per generated level, with approximately 430 re-generations required due to trivial generations.
This is slightly slower than the dataset condition, as some of the levels in the dataset have entities which go outside the bounds of the added border walls, leading to more cases which require rerunning the algorithm.
Notably, the number of trivial generations increased as the partial specification became more constrained.
This could be potentially circumvented with a more robust generation strategy, but for simplicity we chose to simply regenerate in this paper.

As discussed further in the Results section, we found that the dataset setting was the most effective of the three at generating a diverse population of levels.
As such, we present a comparison of that dataset setting against our baselines, and provide the results of the other two settings as an ablation. 

\section{Results}
\begin{figure*}
    \centering
    \includegraphics[width=\linewidth]{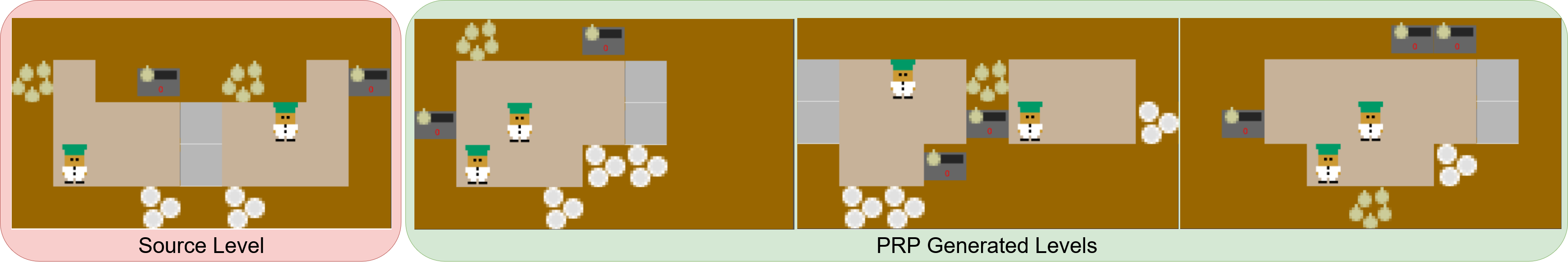}
    \caption{\textit{OvercookedAI} levels generated with PRP. Source level is a human-authored level used as the sole entry in the dataset.}
    \label{overcooked-figure}
\end{figure*}

\begin{figure*}
    \centering
    \includegraphics[width=0.70\linewidth]{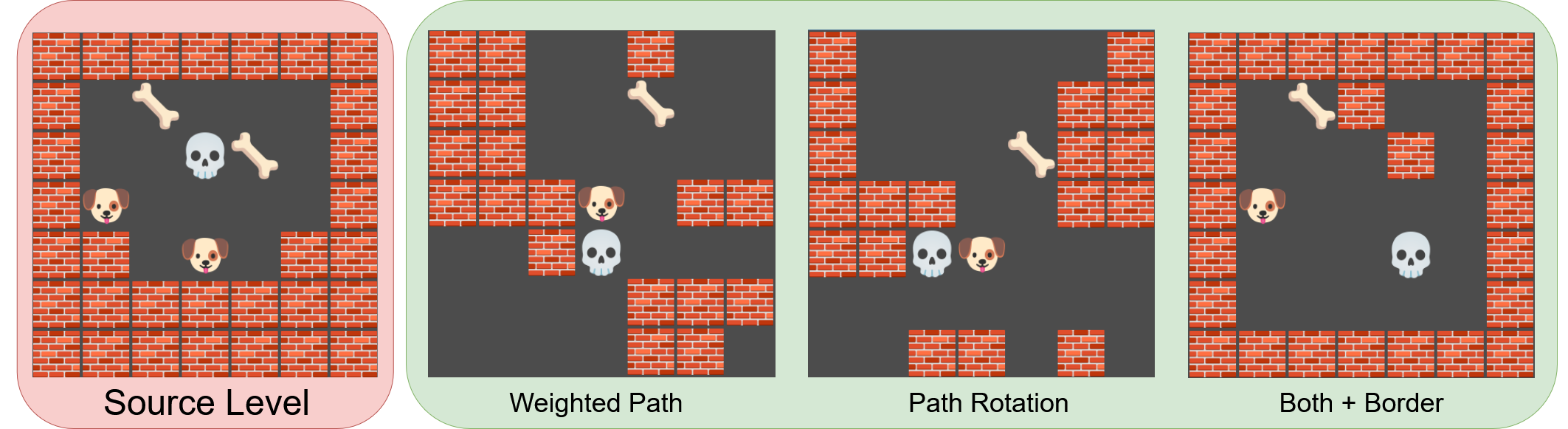}
    \caption{\textit{SkeleWalker} levels generated with PRP. Source level is a human-authored level used as the sole entry in the dataset. Labels under generated levels denote additional settings used in generation.}
    \label{skelewalker-figure}
\end{figure*}

Table \ref{results-table} depicts the comparison of the populations of generated levels as captured by our metrics.
PRP achieves 100\% playability without hand-authored information about \textit{Sokoban}'s dynamics. 
PoD, Sturgeon-MKIII, and Sturgeon-ST also achieve 100\% playability, which is intuitive as each approach requires the output level to be playable via an external check or global constraints.
PCGRL has no such hard requirement and still manages to perform reasonably well in playability, as the approach is guided by a reward function which has multiple hand-authored checks for playability encoded into it \cite{khalifa2020pcgrl}.
The LSTM has no explicit or implicit guidance towards playability, and performs worse on the metric as a result, with over a quarter of the approach's generated levels being unplayable on average.
This demonstrates that any of the common associated approaches for encoding solvability information can be effective in level generation for \textit{Sokoban}. 
However, we specifically identify that PRP accomplishes this without any of these prior common approaches, simply via the cake representation, selection policy, and matching function.

All of the approaches excluding PoD and Sturgeon-ST exhibited very few or no duplicate levels within the generated set.
We hypothesize that PoD's high number of duplicates comes from a convergence towards a singular playable level, which could guide the generation process down the same path of edits multiple times.
Similarly, the additional constraints implemented for Sturgeon-ST's generation process may be causing the approach to have a smaller possibility space for generation, leading to some duplicate generated levels.

PRP exhibited a high level of tile diversity while also retaining a similar amount of unique solutions and signatures to the best performing baselines.
Notably, higher tile diversity is not necessarily always desirable, as a high percentage of tile diversity may indicate the population of levels is noisy.
This is observable when considering the tile diversity of the human dataset, which is lower than PRP, PCGRL Turtle, or the LSTM due to the consistent presence of certain level features such as border walls.

Sturgeon outperforms PRP in unique solution count, but both are similar to the human dataset.
We remind the reader that the Sturgeon levels were $8 \times 8$, which could be a contributing factor.
In terms of unique signatures, all the generated approaches fall short in comparison to the human-authored levels.
PCGRL performs the closest, but we anticipate this may have been due to its reward function biasing the generator towards levels with longer solutions, increasing the chance of more unique signatures. 
Despite no hand-authored factor explicitly biasing it towards unique signatures, PRP still performs similarly to PCGRL, especially in comparison to the other baselines.

We do not provide statistical analysis of our results, as they were not appropriate for these metrics, which are simply sums.
We also lack access to the original generated levels from Zakaria et al.'s paper, making statistical analysis impossible for those populations of levels.
While we do not make any claims that PRP is superior to the existing baseline approaches, it is notable that our approach is able to even perform comparably without the use of external hand-authored signals encoding dynamic information.
Instead, we are able to draw all of the dynamic information from the cake representation, selection policy, and matching function.

Table \ref{ablation-table} depicts the results of our ablation study across our partial specification settings.
Overall, the dataset condition in which the partial specification was empty was most effective in terms of solution and signature diversity.
We hypothesize the decrease in diversity within the other two conditions is due to the constraining nature of a partial specification.
Even by just providing a path or a ring of walls to surround a level, we are drastically reducing the possibility space of potentially generatable levels.
With this in mind, the decreased diversity of levels in the path setting and especially the border setting could be seen as indicating a high degree of controllability.
Guaranteeing existing structure or a specific solution path may be desirable in many applications, such as in co-creative tools \cite{guzdial2018co}. 


\section{Case Studies}

To showcase the generalizability of PRP, we present two additional case studies in two separate game domains. 
While we do not provide robust evaluations for these case studies, they instead represent PRP's applicability as a generation approach to multiple domains with minimal domain-specific changes. 
The two domains were \textit{OvercookedAI}, an existing grid-world domain commonly used in reinforcement learning research \cite{carroll2019utility}, and \textit{SkeleWalker}, a proof of concept domain we authored.
These domains showcase PRP's applicability to domains with more complex entities and rules, and PRP's potential usefulness as part of the design process respectively.

\subsection{\textit{OvercookedAI}}

\textit{OvercookedAI}\footnote{https://github.com/HumanCompatibleAI/overcooked\_ai} is a simplification of the game \textit{Overcooked}. 
In \textit{OverCookedAI}, two chefs, controlled by humans or AI agents, work together to prepare meals.
A defining feature of the environment is the complexity of the paths required to achieve reward.
Agents must go through several steps to cook various recipes before delivering food to a serving location.
In particular, agents must collect onions, place them on the stove, wait for the onions to cook, collect a plate, place the cooked onion on the plate, then deliver them to the serving location.
There is additional complexity added with the multiplayer aspect of the environment.
Players can pass objects over walls to each other, which is sometimes required to achieve reward in human-authored levels.
As such, it is challenging to generate valid environments for \textit{OvercookedAI} which are non-trivial and distinct from existing training data \cite{rocha2025procedural}.

Figure \ref{overcooked-figure} depicts three randomly selected generated levels based on a single input level and playtrace, which contains movement information for both players.
Our procedure for generating these levels was similar to our \textit{Sokoban} implementation, with some tweaks based on the domain.
As in \textit{Sokoban}, \textit{OvercookedAI} is on a grid, so we perform a similar tile to character conversion to process the levels to and from the cake representation.
For the PRP selection policy, we first considered the two player paths.
We then consider all other objects simultaneously, which are implicitly restricted by the places where the player paths interacted with each object in the cake representation.
The matching function was implemented similarly to \textit{Sokoban}, allowing only two player paths and for entities to not overlap on the same timestep.
After generation is complete, we perform the same simple preprocessing as in \textit{Sokoban}, taking the first slice of the output cake as the output level.

As shown in Figure \ref{overcooked-figure}, PRP generated meaningfully distinct, playable environments from a single example.
In the source level, the two players are separated by serving plates and must work independently.
However, in the presented generated environments, there are examples where the two players are together or with different objects separating the players.
This leads to different optimal solutions and potential strategies, which could be beneficial for the training and evaluation of reinforcement learning agents in the domain \cite{ruhdorfer2024overcooked, fontaine2021importance}.
However we note in all cases the original player paths are still valid solutions.
While we do not present a robust evaluation of PRP in this domain, it is a promising showcase of the generalizability of our approach to more complex domains.

\subsection{\textit{SkeleWalker}}
\textit{SkeleWalker} is a test domain we designed to learn more about the generalizability of the cake representation and PRP, as a variant of \textit{Sokoban}.
In the game, the player controls a skeleton tasked with the goal of walking each dog to a stationary bone, as opposed to how the player pushes boxes into goals in \textit{Sokoban}.
If the skeleton is adjacent to any one dog, the dog begins to follow the skeleton, tracing its path with every movement action.
If the skeleton is ever followed by more than one dog, the skeleton gets eaten, and the game is considered lost.
The game is considered won when all dogs are occupied by a bone.

We designed \textit{SkeleWalker} to test the PRP algorithm in a unique domain, and to see how well the approach would generalize with as few changes as possible from the \textit{Sokoban} implementation.
As \textit{SkeleWalker} is also a grid-based game with similar rules and entities to \textit{Sokoban}, we used the exact same selection policy and representation, with dogs in place of boxes, and attempted to generate levels just as in \textit{Sokoban}.
However, we found that PRP was exactly replicating the dog placements found in the original levels with only translational variations. 
We hypothesize this is due to the more constrained nature of the paths in \textit{SkeleWalker}, as the player needs to walk directly over each goal rather than just pushing a box to it and the gameplay has additional constraints around relative placements of dogs.
In other words, there were fewer valid levels that could be output following the PRP assumptions for this domain with a single source level.
To get around this, we implemented a weighted path augmented similar to the one found in the ``path'' setting of our ablation test where we provided a player path as a partial specification.
In particular, this weighted path was generated such that at each timestep, the path had a 50\% chance to continue in its current direction, and a 50\% chance to pick a new direction.
This choice was made given the nature of the paths required to make unique \textit{SkeleWalker} levels, and gives some insight into the PRP algorithm: not all input playtraces are necessarily good for generating a variety of levels, particularly when working from a single example.
Further, we introduced path rotation for the dog paths in the MatchesAt function, which acted as a direct transformation to the pool of possible temporal entities able to be matched to a given player path.

Figure \ref{skelewalker-figure} depicts three generated \textit{SkeleWalker} levels with varying conditions, as well as the source level they were based on.
Notably, while the positions of the dogs relative to the bones are the same across levels, each level has a different optimal solution path due to the initial position of the player.
We hypothesize that with more input levels, we would see a larger variety in potential level possibilities, as we found in the  \textit{Sokoban} domain.
In particular, levels with shorter temporal entity paths in theory create a higher amount of possible variety in output levels, as there are simply more places in time and space where entities can be placed relative to other entities already present in the output cake.
Notably, the cake representation was not changed between \textit{Sokoban} and \textit{SkeleWalker}, despite encoding a completely different set of mechanics.
The only changes that were made between the two domains were to the PRP settings previously discussed.
We feel these adaptations required to make the algorithm perform in the \textit{SkeleWalker} domain highlight the strengths and weaknesses of PRP, and showcases the versatility of the cake representation.

\section{Discussion}
In this section, we discuss the limitations of the cake representation, PRP, and our evaluation.
We also discuss future avenues for exploration for the cake representation and PRP.

\subsection{Limitations}
One limitation of the cake representation is the requirement of domain-specific processing from and to the initial game's representation.
This requirement is shared with similar text-based tile representations \cite{summerville2016vglc}. 
Sometimes this required information may involve the game's dynamics, meaning the processing must have additional dynamic information about a game's systems.
However, this is a trade-off in terms of where the domain-specific information is encoded.
By only requiring domain-specific information during processing to and from the cake representation, we argue that this allows algorithms that use the cake representation to be more domain-independent. Notably in all three domains presented in this paper, the base PRP algorithm did not change.

Another limitation is that the conversion from the original game's representation to the cake representation may be lossy.
The amount of information lost in the conversion is dependent on the chosen representation for each level slice in the cake representation.
For example, in \textit{Super Mario Bros.}, Mario's movement is not bound by the tile grid of blocks that make up the level geometry.
As such, potentially important information could be lost when converting the gameplay into a cake-style representation using a tile grid.
This could be possible to avoid with tweaks to how information for each slice of cake is represented, such as appending Mario's relative x and y position to each cake slice rather than representing him as a single token.
However, this requires additional insight into the game's dynamics, and an approach using this adjusted representation would also need to be updated on a per-game basis.
As such, we present a more traditional tile-based representation in this paper for simplicity, and leave further adaptation to future work.

A limitation of our evaluation is the use of some existing results and generated level corpora rather than reimplementing and regenerating new populations of levels for all baselines.
While we acknowledge this limitation, this paper is not intended to be a broad evaluation in the domain of \textit{Sokoban}.
Rather, we showcase a new generation approach using our cake representation and intend to show it can perform comparably to other existing approaches using additional hand-authored information.

\subsection{Future Work}
There are a number of future research avenues for both the cake representation and PRP.

For the cake representation, one could try this representation with existing level generation approaches.
Constraint-based systems such as Sturgeon or WFC are a natural fit for the cake representation \cite{cooper2022sturgeon,vandara2025spacetime}.
As presented in this paper, the cake representation is extendable to other 2D grid-based games.
In terms of changing the representation, there is a valuable extension of the cake representation to environments which are not grid based, such as 3D games.
This could involve taking a similar approach of representing the level state over time in slices, but the representation of each level slice would need to change.
Possible solutions could be polygon-based or graph-based representations of these levels.


For the PRP approach, there are a number of potential application domains.
PRP could be applicable to many reinforcement learning tasks, in which agents are trained to accomplish a goal based on their interactions with an environment.
PRP is especially suitable to these tasks, as we can generate many variations of levels based on an agent's playtrace that are consistently valid.
These generated levels could be useful in the training process, as well as to evaluate the behaviour of existing agents.
We also imagine PRP could serve as the basis for co-creative game design tools.
As an example of a potential tool, designers could create a partial specification with an editor, then browse levels generated by PRP in a search-based format, picking and choosing the levels they like best.

We have preliminary results of the application of PRP to the 3D grid-based domain \textit{Minecraft} using an altered representation to allow for the encoding of multiple playthroughs with branching edits.
We intend to present these results in future work.

\section{Conclusions}
In this paper, we introduced the cake representation, a novel structure for representing game levels over time.
We introduced PRP, a generation approach designed to use the cake representation, and found that PRP performed comparably to state-of-the-art baselines with additional human-authored information in the domain of \textit{Sokoban}.
We consider the cake representation a promising step toward addressing the issue of dynamics in level representation, and hope it can prove useful for future technical games research.

\begin{acks}
This work was funded by the Canada CIFAR AI Chairs Program. We acknowledge the support of the Alberta Machine Intelligence Institute (Amii). The authors would like to thank Kaylah Facey and Seth Cooper for helpful discussions.
\end{acks}

\bibliography{main}

\end{document}